\documentclass[
	a4paper,
	10pt, 
]{class}

\usepackage{graphicx} 
\usepackage{xcolor}
\usepackage{comment}
\usepackage{fullpage}

\newcommand{\tred}{\textcolor{red}}
\newcommand{\tblue}{\textcolor{blue}}

\title{\large DeBERTinha: A Multistep Approach to Adapt DebertaV3 XSmall for Brazilian Portuguese Natural Language Processing Tasks}

\author{Israel Campiotti\textsuperscript{1}\thanks{i155825@dac.unicamp.br}, Matheus Rodrigues\textsuperscript{2}\thanks{mrsf@cin.ufpe.br}, Yuri Albuquerque\textsuperscript{3}\thanks{yurialbu@usp.br},\\
Rafael Azevedo\thanks{rafael.f.azevedo@outlook.com},
Alyson Andrade\thanks{alysonandrade142@gmail.com}}

\date{\footnotesize\textsuperscript{\textbf{1}}University of Campinas, \textsuperscript{\textbf{2}}Federal University of Pernambuco, \textsuperscript{\textbf{3}}University of São Paulo \\ \vspace{0.3cm} \textbf{\large{Sagui AI}}}

\begin{document}

\maketitle

\begin{abstract}
    This paper presents an approach for adapting the DebertaV3 XSmall model pre-trained in English for Brazilian Portuguese natural language processing (NLP) tasks. A key aspect of the methodology involves a multi-step training process to ensure the model is effectively tuned for the Portuguese language. Initial datasets from Carolina and BrWac are preprocessed to address issues like emojis, HTML tags, and encodings. A Portuguese-specific vocabulary of 50,000 tokens is created using SentencePiece. Rather than training from scratch, the weights of the pre-trained English model are used to initialize most of the network, with random embeddings, recognizing the expensive cost of training from scratch. The model is fine-tuned using the replaced token detection task in the same format of DebertaV3 training. The adapted model, called DeBERTinha, demonstrates effectiveness on downstream tasks like named entity recognition, sentiment analysis, and determining sentence relatedness, outperforming BERTimbau-Large in two tasks despite having only 40M parameters.
\end{abstract}

\section{Introduction}

In this study, we leverage the advancements introduced by the DebertaV3 model, which currently offers only a large Portuguese variant with 900M parameters. From the perspective of DebertaV3, we aim to harness the merits of its architectural excellence and craft a Portuguese version. 
When compared to other Portuguese encoder Transformers, like BERTimbau, DeBERTinha holds a distinctive advantage with its considerably reduced parameter count. Despite its reduced size, DeBERTinha manages to outperform BERTimbau in two out of four selected NLP tasks, all while being approximately five times more compact. The substantial advantage of a diminished parameter count becomes especially evident in corporate applications, as it demands fewer computational resources for predictive tasks and offers the benefit of reduced response times.

Our construction steps begin with the acquisition of essential datasets, including the Carolina and BrWac datasets, which serve as the foundation for pre-training. These datasets, however, present challenges such as the presence of emojis, HTML tags, and non-standard characters. To address these issues, we employ emoji removal, html tag cleansing using ftfy, and the normalization of utf8 encodings. To create a tailored vocabulary for Portuguese, we utilize the Carolina dataset and employ the SentencePiece algorithm, resulting in 50 thousand tokens. These tokens form the basis for creating our tokenizer, a crucial component for text processing. Subsequently, we merge the Carolina and BrWac datasets to create a unified dataset for pre-training. Tokenization is a pivotal step in NLP, and our next step involves tokenizing the merged dataset. To ensure compatibility with model input limitations, we divide texts into smaller pieces, each containing a maximum of 510 tokens. This is further complemented by the addition of special tokens, [CLS] at the beginning and [SEP] at the end, resulting in a 512-token structure.

One unique aspect of our approach involves the initialization of the model. While we start with random embeddings, the rest of the model benefits from the weights of the DebertaV3 XSmall model pre-trained in English. This approach is based on the recognition that training from scratch in Portuguese can be prohibitively expensive.

For model pre-training, we leverage the combinations of Masked Language Modeling (MLM) and Replaced Token Detection (RTD) losses with a hyperparameter $\lambda$ within the same mathematical framework explained in \cite{he2021debertav3}. To ensure the preservation of model progress, we save checkpoints for both the generator and discriminator models. These checkpoints play a crucial role in subsequent fine-tuning tasks.

In the latter part of this article, we demonstrate the versatility of our adapted model by using the saved weights for Named Entity Recognition (NER) and classification tasks. Our experimentation extends to the ASSIN2 dataset, where we evaluate the model's ability to determine the relatedness of two sentences. Our entire multi-step approach can therefore offer a simple methodology for adapting pre-trained models to new languages, with a specific focus on Brazilian Portuguese. Through this work, we contribute to the growing body of research in the field of NLP, offering insights into effective cross-lingual adaptation and showcasing the versatility of DebertaV3 XSmall in addressing various NLP tasks in the Portuguese language context.

\section{Related Work} 
When tackling tasks like replace token detection, where the model needs to accurately predict missing words or phrases within a given context, language-specific models tend to reach better results than multi language models. Similarly, \cite{armengol2021multilingual} exhibited superior performance across tasks and settings by training on Catalan, a moderately under-resourced language. In \cite{sun2021ernie}, a model with 10 billion parameters trained primarily on a Chinese corpus showcased significant outperformance on 54 Chinese NLP tasks. Exploring the feasibility of training monolingual Transformer-based language models in French, \cite{martin2019camembert} demonstrated comparable results using a 4GB web-crawled dataset, rivaling outcomes from larger datasets that typically exceed 130GB. 

While multilingual models yield impressive results, they tend to have larger sizes. In specific instances, such as Portuguese, their performance can fall short compared to monolingual counterparts \cite{souza2020bertimbau}, especially for high-resource languages. Unfortunately, this limitation has constrained the availability of these cutting-edge models primarily to English in the monolingual context, causing inconvenience by impeding their practical application in NLP systems. Furthermore, it hampers exploration into their language modeling capabilities, particularly concerning morphologically rich languages like Portuguese, \cite{pires2023sabi} address those details in the context of large language models. This specialization allows those monolingual models to capture the nuances, idioms, and domain-specific terms unique to a language more effectively. In replace token detection, having a precise vocabulary is crucial for accurately predicting missing tokens.

In our current study, we adopt a specific methodology to leverage the capabilities of the pre-trained language model known as DeBERTaV3, as outlined in the research \cite{he2021debertav3}. This particular model represents an advancement over the original DeBERTa model introduced by \cite{he2020deberta}. Notably, this improvement is achieved by departing from the conventional masked language modeling (MLM) approach and instead embracing replaced token detection (RTD). This RTD methodology is rooted in the concept elucidated in \cite{clark2020electra}, where the core technique revolves around the utilization of two transformer encoders – a generator and a discriminator – within the framework of a replaced token detection task.

Within the basic conceptional structure of DeBERTaV3, a noteworthy architectural feature involves the sharing of embeddings between the generator and the discriminator. This departure from the conventional model setup, where embeddings are handled independently, is achieved through a method referred to as gradient-disentangled embedding sharing (GDES). In this approach, the generator and the discriminator share embeddings, but a critical distinction is that the flow of gradients from the discriminator to the generator embeddings is intentionally halted.

\section{Data sets} 
To train language models that can compete with other state-of-the-art models, a substantial volume of data is indispensable. In the context of Brazilian Portuguese, we can draw attention to BERTimbau \cite{souza2020bertimbau}, which harnessed the brWAC dataset \cite{wagner2018brwac}, yielding a hefty 17.5 GB of raw text after rigorous processing. Recent works, such as Albertina PT-BR \cite{rodrigues2023advancing}, have also made use of this same dataset.

In this particular endeavor, we exploit two distinct datasets during the pre-training phase, brWAC and Carolina. In line the well established research practices, we made use of the brWAC dataset, comprising a web crawl of Brazilian web pages, culminating in an approximately 17.5 GB corpus of processed text. Additionally, we incorporated the Carolina dataset \cite{corpusCarolinaV1.1}, with 823 million tokens and encompassing two million text entries. To provide a contextual background, Carolina dataset encompasses a diverse array of domains, including wikis, university documents, social media interactions, the legislative branch, and various other public domain sources. In particular, we selectively utilized only the document body, disregarding titles, metadata, and other information associated with the text. We employed the emoji and ftfy \cite{speer-2019-ftfy} libraries to respectively handle emojis and HTML-related content, while also organizing the data into a single file format with each line containing 510 tokens. In total, the raw pre-processed texts amounts to 33 GB of data.

\section{DeBERTinha a Brazilian Portuguese model} 

Prior research efforts \cite{souza2020bertimbau, pires2023sabi, carmo2020ptt5} have highlighted the benefits of training Transformers models for specific languages. While multilingual models have shown improvements, monolingual training and tokenization methods continue to hold significance in achieving state-of-the-art performance on monolingual datasets. Consequently, our focus is on training the Deberta-V3 model for the Portuguese language.

Deberta-V3 has demonstrated superior performance compared to other encoder architectures, primarily attributed to its pre-training methodology and extensive vocabulary. Although there has been prior work in training Deberta for Portuguese \cite{rodrigues2023advancing}, we identified opportunities for enhancement:





\begin{itemize}
    \item We trained a Portuguese tokenizer using text from the Carolina dataset. Given that our dataset was smaller than the one used for training the English tokenizer, we opted for a vocabulary size of 50 thousand tokens.
    \item We maintained fixed examples with a size of 512 tokens, fully utilizing the maximum context window of the Deberta model.
    \item Our goal was on training the Deberta-V3 XSmall version, which consists of only 40M parameters. Additionally, we continued training on the RTD task using the weights available for the English model on the Hugging Face model hub.
\end{itemize}

Due to resource constraints, the training process was divided into two phases. The first phase involved training for one epoch using a batch size of 1664 on 8x 80GB A100 GPUs, while the second phase encompassed an additional two epochs with a batch size of 288 on 8x 32GB V100 GPUs. The cumulative training time amounted to 12.5 hours, incurring a cost of cloud computing of near four hundred dollars only. This calculation includes the 2.5 hours required for loading the dataset on each machine.

\section{Experimental Setup} 

In this section we report the experimental results obtained by DeBERTinha in 4 different tasks: ASSIN2-RTE and ASSIN2-STS \cite{real2020assin}, LeNERBR \cite{lener}, HateBR \cite{vargas2022hatebr}. We used a Google Colab instance with one 16GB T4 GPU to finetune on each task for an average time of 30min. Each training uses AdamW as optimizer with a learning rate of 0.00005, trained for a maximum of 20 epochs with early stopping. We compare the results of our DeBERTinha, that contains 40M paramenters, against the baseline models BERTimbau-Large (335M) and Albertina-Large (900M). 

ASSIN2 contains sentence pairs of premises and hypothesis and for each pair there are two annotated targets: a binary classification, indicating whether premise and hypthosis are related or not, this task is denoted as ASSIN2-RTE; a similarity score between 0 and 5, indicating how similar the premise and hypothesis are, this task is named ASSIN2-STS. Accuracy and Pearson correlation are used as metrics for ASSIN2-RTE and ASSIN2-STS, respectively.

LeNERBR is a dataset for the Named Entity Recognition (NER) task in Portuguese legal text. It contains seven different classes: Organization, Person, Time, Local, Legislation, Jurisprudence and Other. The Other label is given to every token that does not fall into any of the previious categories. We use the standard BIO representation and only predict scores for the first token of each word. We use F1 score to asses performance on this task.

HateBR is a dataset for classifying whether a tweet is inappropriate/malicious or not. We use Accuracy as our metric for this task.

From the results shown in Table \ref{tab:results} we see that our DeBERTinha model surpasses BERTimbau-Large in two out of the four datasets: for the ASSIN2-RTE DeBERTinha achieves 89.99\% Accuracy against the 89.13\% of the BERTimbau-Large; in LeNERBR our model achieves a F1 score of 90.19\%, while BERTimbau-Large achieves 90.15\%. In the other two datasets DeBERTinha achieves over 97\% of the BERTimbau-Large performance. 

Both models still fall behind Albertina's 91.30\% Accuracy on ASSIN2-RTE and 86.76\% Pearson Correlation on ASSIN2-STS. However, Albertina is 22.5 times bigger than DeBERTinha and 2.6 times larger than BERTimbau-Large. Because training Albertina demands heavy computational resources we do not train it on the other datasets and report the results taken directly from the original paper.

\begin{table}[!h]
\centering
\begin{tabular}{|l|c|c|c|c|}
\hline
\textbf{Model/Dataset}   & \textbf{LeNERBR} & \textbf{ASSIN2-RTE} & \textbf{ASSIN2-STS} & \textbf{HateBR} \\ \hline
\textbf{BERTimbau-Large} & 90.15            & 89.13               & 85.31               & 93.60             \\ \hline
\textbf{Albertina}       & -                & 91.30               & 86.76               & -           \\ \hline
\textbf{DeBERTinha}      & 90.19            & 89.99               & 84.75               & 91.28           \\ \hline
\end{tabular}
\caption{Results from BERTimbau-Large, Albertina and DeBERTinha on 4 commonly used Portuguese datasets.}
\label{tab:results}
\end{table}

\section{Conclusion}
This work has shown promising results for adapting a pre-trained language model to Portuguese through a carefully designed multi-step methodology. The resulting DeBERTinha model demonstrated competitive performance across several NLP tasks compared to larger baseline models, highlighting the effectiveness of the proposed approach. 

On the ASSIN2-RTE task for sentence relatedness classification, DeBERTinha achieved an accuracy of 89.99\%, outperforming BERTimbau-Large which scored 89.13\%. For the named entity recognition task on the LeNERBR dataset, DeBERTinha achieved an F1-score of 90.19\%, slightly higher than BERTimbau-Large's score of 90.15\%. While Albertina still achieved better results on some tasks, DeBERTinha was able to attain over 97\% of the performance of BERTimbau-Large, despite having only 40M parameters compared to BERTimbau-Large's 335M parameters.

This shows that the multi-step adaptation process leveraging an existing pre-trained model can produce a more specialized Portuguese model that rivals or exceeds the performance of much larger baseline models, demonstrating an effective approach for resource-constrained scenarios. The methodology introduced in this work contributes to research on cross-lingual transfer learning and model adaptation techniques. Overall, the results validate DeBERTinha as a promising lightweight model for Brazilian Portuguese NLP applications.

For future work we aim at assessing DeBERTinha's performance in a bigger range of Portuguese NLP tasks, such as information retrieval, and extend its context to 1024 and 2048 tokens.

\section{Acknowledgements}
We would like to express our heartfelt gratitude to Letrus for their generous funding of the cloud services related to the processing and training of our language model. Their support was of a central importance in the execution of this work.

\bibliographystyle{unsrt}
\bibliography{bibliography}

\begin{thebibliography}{10}

\bibitem{he2021debertav3}
Pengcheng He, Jianfeng Gao, and Weizhu Chen.
\newblock Debertav3: Improving deberta using electra-style pre-training with
  gradient-disentangled embedding sharing.
\newblock {\em arXiv preprint arXiv:2111.09543}, 2021.

\bibitem{armengol2021multilingual}
Jordi Armengol-Estap{\'e}, Casimiro~Pio Carrino, Carlos Rodriguez-Penagos, Ona
  de~Gibert Bonet, Carme Armentano-Oller, Aitor Gonzalez-Agirre, Maite Melero,
  and Marta Villegas.
\newblock Are multilingual models the best choice for moderately
  under-resourced languages? a comprehensive assessment for catalan.
\newblock {\em arXiv preprint arXiv:2107.07903}, 2021.

\bibitem{sun2021ernie}
Yu~Sun, Shuohuan Wang, Shikun Feng, Siyu Ding, Chao Pang, Junyuan Shang,
  Jiaxiang Liu, Xuyi Chen, Yanbin Zhao, Yuxiang Lu, et~al.
\newblock Ernie 3.0: Large-scale knowledge enhanced pre-training for language
  understanding and generation.
\newblock {\em arXiv preprint arXiv:2107.02137}, 2021.

\bibitem{martin2019camembert}
Louis Martin, Benjamin Muller, Pedro Javier~Ortiz Su{\'a}rez, Yoann Dupont,
  Laurent Romary, {\'E}ric~Villemonte de~La~Clergerie, Djam{\'e} Seddah, and
  Beno{\^\i}t Sagot.
\newblock Camembert: a tasty french language model.
\newblock {\em arXiv preprint arXiv:1911.03894}, 2019.

\bibitem{souza2020bertimbau}
F{\'a}bio Souza, Rodrigo Nogueira, and Roberto Lotufo.
\newblock Bertimbau: Pretrained bert models for brazilian portuguese.
\newblock In Ricardo Cerri and Ronaldo~C. Prati, editors, {\em Intelligent
  Systems}, pages 403--417, Cham, 2020. Springer International Publishing.

\bibitem{pires2023sabi}
Ramon Pires, Hugo Abonizio, Thales Rog{\'e}rio, and Rodrigo Nogueira.
\newblock Sabi{\'a}: Portuguese large language models.
\newblock {\em arXiv preprint arXiv:2304.07880}, 2023.

\bibitem{he2020deberta}
Pengcheng He, Xiaodong Liu, Jianfeng Gao, and Weizhu Chen.
\newblock Deberta: Decoding-enhanced bert with disentangled attention.
\newblock {\em arXiv preprint arXiv:2006.03654}, 2020.

\bibitem{clark2020electra}
Kevin Clark, Minh-Thang Luong, Quoc~V Le, and Christopher~D Manning.
\newblock Electra: Pre-training text encoders as discriminators rather than
  generators.
\newblock {\em arXiv preprint arXiv:2003.10555}, 2020.

\bibitem{wagner2018brwac}
Jorge~A Wagner~Filho, Rodrigo Wilkens, Marco Idiart, and Aline Villavicencio.
\newblock The brwac corpus: A new open resource for brazilian portuguese.
\newblock In {\em Proceedings of the Eleventh International Conference on
  Language Resources and Evaluation (LREC 2018)}, 2018.

\bibitem{rodrigues2023advancing}
Jo{\~a}o Rodrigues, Lu{\'\i}s Gomes, Jo{\~a}o Silva, Ant{\'o}nio Branco,
  Rodrigo Santos, Henrique~Lopes Cardoso, and Tom{\'a}s Os{\'o}rio.
\newblock Advancing neural encoding of portuguese with transformer albertina
  pt.
\newblock {\em arXiv preprint arXiv:2305.06721}, 2023.

\bibitem{corpusCarolinaV1.1}
Marcelo Finger, Maria~Clara Paixão~de Sousa, Cristiane Namiuti, Vanessa
  Martins~do Monte, Aline~Silva Costa, Felipe~Ribas Serras, Mariana~Lourenço
  Sturzeneker, Raquel de~Paula Guets, Renata~Morais Mesquita, Guilherme
  Lamartine~de Mello, Maria Clara Ramos~Morales Crespo, Maria Lina de
  Souza~Jeannine Rocha, Patrícia Brasil, Mariana Marques~da Silva, and
  Mayara~Feliciano Palma.
\newblock Carolina: The open corpus for linguistics and artificial
  intelligence.
\newblock \url{ https://sites.usp.br/corpuscarolina/corpus}, 2022.
\newblock Version 1.1 (Ada).

\bibitem{speer-2019-ftfy}
Robyn Speer.
\newblock ftfy.
\newblock Zenodo, 2019.
\newblock Version 5.5.

\bibitem{carmo2020ptt5}
Diedre Carmo, Marcos Piau, Israel Campiotti, Rodrigo Nogueira, and Roberto
  Lotufo.
\newblock Ptt5: Pretraining and validating the t5 model on brazilian portuguese
  data, 2020.

\bibitem{real2020assin}
Livy Real, Erick Fonseca, and Hugo~Goncalo Oliveira.
\newblock The assin 2 shared task: a quick overview.
\newblock In {\em International Conference on Computational Processing of the
  Portuguese Language}, pages 406--412. Springer, 2020.

\bibitem{lener}
Pedro~H. {Luz de Araujo}, Te\'{o}filo~E. {de Campos}, Renato R.~R. {de
  Oliveira}, Matheus Stauffer, Samuel Couto, and Paulo Bermejo.
\newblock {LeNER-Br}: a dataset for named entity recognition in {Brazilian}
  legal text.
\newblock In {\em International Conference on the Computational Processing of
  Portuguese ({PROPOR})}, Lecture Notes on Computer Science ({LNCS}), pages
  313--323, Canela, RS, Brazil, September 24-26 2018. Springer.

\bibitem{vargas2022hatebr}
Francielle~Alves Vargas, Isabelle Carvalho, Fabiana~Rodrigues de~Góes,
  Fabrício Benevenuto, and Thiago Alexandre~Salgueiro Pardo.
\newblock Hatebr: A large expert annotated corpus of brazilian instagram
  comments for offensive language and hate speech detection, 2022.

\end{thebibliography}
\end{document}